\documentclass{article}
\usepackage{spconf,amsmath,graphicx,tabularx,booktabs,multirow,svg}
\usepackage{hyperref}

\title{Unsupervised Extractive Dialogue Summarization\\in Hyperdimensional Space}
\name{Seongmin Park, Kyungho Kim, Jaejin Seo, Jihwa Lee}
\address{ActionPower\\
Seoul, Republic of Korea}
\begin{document}
%
\maketitle
\begin{abstract}
We present HyperSum, an extractive summarization framework that captures both the efficiency of traditional lexical summarization and the accuracy of contemporary neural approaches. HyperSum exploits the pseudo-orthogonality that emerges when randomly initializing vectors at extremely high dimensions ("blessing of dimensionality") to construct representative and efficient sentence embeddings. Simply clustering the obtained embeddings and extracting their medoids yields competitive summaries. HyperSum often outperforms state-of-the-art summarizers -- in terms of both summary accuracy and faithfulness -- while being 10 to 100 times faster. We open-source HyperSum as a strong baseline for unsupervised extractive summarization\footnote{\url{https://github.com/seongminp/hyperseg}}.
\end{abstract}
\begin{keywords}
Hyperdimensional computing, vector symbolic architectures, summarization
\end{keywords}

\section{Introduction}
\label{sec:intro}

Modern extractive summarizers are steadily yielding the spotlight to their abstractive counterparts, whose barriers to usage have been significantly lowered by improvements in efficient neural language modeling. Notably, research on extractive summarization increasingly focuses on incorporating extractive summarizers as preliminary summarization components \cite{Xiao_Wang_He_Jin_2020, ijcai2022p0764} or context reducers \cite{bishop-etal-2022-gencomparesum} in hybrid language processing pipelines. Therefore, the efficiency and accuracy of modern extractive summarizers have become more crucial, as any delays or inaccuracies will propagate downstream through the pipeline. With these considerations, we propose HyperSum, a new extractive summarizer that captures both priorities.

\begin{figure}[t]
\centering
\includegraphics[width=0.9\columnwidth]{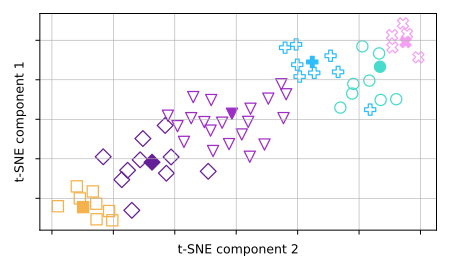}
\caption{HyperSum's utterance embeddings for clip \#6 from the Behance dataset, visualized with t-SNE \cite{van2008visualizing}. Different shapes denote different sentence clusters. Shaded markers in each cluster are medoids, which are selected as its representative summary.}
\label{fig:overview}
\end{figure}

HyperSum is the first attempt to integrate recent inspirations in hyperdimensional computing (HDC) \cite{kanerva2009hyperdimensional, gorban2018blessing} into text summarization. HDC enables extremely efficient but representative initialization of sequence embeddings, a trait beholden to emergent pseudo-orthogonality of random vectors in high dimensions. In our experiments, we find that such representational capabilities of HDC synergetically support the aforementioned performance requirements of modern extractive summarization. High-quality sentence embeddings allow the pairing of uncomplicated and efficient extractive summarization schemes, such as simple medoid selection (Figure \ref{fig:overview}). On a CPU, HyperSum reduces summarization runtime by orders of magnitudes compared to neural summarizers operating on GPUs, while generating more accurate summaries. We outline our contributions as follows:
\begin{enumerate}
    \item We present the first framework to use hyperdimensional computing in text summarization.
    \item HyperSum outperforms previous state-of-the-art extractive summarization systems while being several orders of magnitude more efficient.
    \item We perform extensive ablation studies to examine which configuration and tokenization schemes best utilize the representational advantages offered by HDC.
\end{enumerate}

\section{Background}
\label{sec:background}

\subsection{Hyperdimensional computing}

HDC (also known as vector symbolic architectures) is an emerging class of representation embeddings that lies between symbolic artificial intelligence and distributed semantics. A random vector at high enough dimensions — typically over 10,000 — will be pseudo-orthogonal to any other randomly initialized vector (i.e. probabilistically dissimilar as measured by distance metrics such as cosine distance). This pseudo-orthogonality is used interchangeably with true orthogonality in practical applications such as text classification and on-device machine learning \cite{najafabadi2016hyperdimensional, kleyko2023survey}. This emergent behavior, called the \textit{concentration of measure} \cite{ledoux2001concentration}, sets the stage for creative applications in representation learning. One such example is the ability to create hyperdimensional vectors that basket other vectors and become similar to its every constituent, but remain dissimilar to all other randomly initialized hypervectors. In this work, HyperSum uses the concentration of measure to construct sentence embeddings. While HDC is being gradually introduced to language processing domains \cite{najafabadi2016hyperdimensional, kanerva2010we}, this study is the first to experiment with applying the benefits of HDC to text summarization.    

\subsection{Extractive summarization}

Extractive summarization selects $n$ most representative sentences from a document as its summary. For the scope of this paper, we assume $n$ is known. Extractive summarizers are used both as stand-alone systems and, more recently, as part of hierarchical summarization schemes. The integration of extractive summarizers into hybrid summary pipelines demands the summarizers to be both efficient and accurate. Therefore, we explore a wide range of baseline extractive summarizers that specialize in both spectrums. 

Textrank \cite{mihalcea2004textrank} is a graph-based summarizer that identifies nodes based on lexical centrality. PacSum \cite{zheng-lapata-2019-sentence} also utilizes sentence centrality to identify the most important sentences in a document. Each sentence in the document is embedded with a neural language model and added as a node to a sentence graph. OTExtSum \cite{tang-etal-2022-otextsum} selects a set of summary sentences that minimizes the semantic distance between the candidate summary set and the original document. Following the previous literature, we also include Lead-$n$ \cite{zhu2020make} as our baseline. Lead-$n$ selects the $n$-leading sentences as the summary, exploiting the lead bias typically present in documents. We use BERT variants of PacSum and OTExtSum as our baselines for their state-of-the-art performance.

\section{Methodology}
\label{sec:methodology}
HyperSum consists of two steps: \textit{sentence embedding construction}, in which we build sentence representations at hyper-dimension, and \textit{summary extraction}, where we select the most important sentences among the vectorized sentences. Since HDC efficiently converts each sentence into representative vectors, we simply take the medoid of every sentence cluster as the summary sentence. 

\subsection{Constructing sentence embeddings}
Following \cite{najafabadi2016hyperdimensional}, we build sentence embeddings by bundling hyperdimensional vector embeddings of all words in the sentence. The pseudo-orthogonality of hypervectors allows word embeddings and sentence embeddings to share the same dimension. We assign a random thermometer hypervector \cite{thermometer} to every unique word $\boldsymbol{w_i}$ in the document. 
\begin{align}
\boldsymbol{w_i} = <T(i,0), …, T(i,D-1)>, \\
\text{where } 
T(i,d) = \begin{cases}
-1,\quad d < i\\
\enspace \textit{ }1,\quad d \ge i
\end{cases}. \nonumber
\end{align}
$D$ is the global hypervector dimension. In this paper we use $D=10,000$ for all experiments.
$T$ is the temperature encoding function that sets the value for each index of $\boldsymbol{w_i}$.

In other words, the $D$-dimensional bipolar hypervector for $\boldsymbol{w_i}$ has the value $-1$ at indices under $i$ and $1$ at indices above and including $i$.
Same words in different sentences share the same hyperdimensional embedding. 

The hyperdimensional embedding of each sentence is constructed by "binding" every constituent word embedding \cite{najafabadi2016hyperdimensional, kleyko2023survey}:
To generate a sentence vector $\boldsymbol{u}$, we first encode the position of each constituent word vector $\boldsymbol{w_{j}}$ by right-shifting ($\Pi$) $l-j-1$ times. $l$ is the sentence word count and $j$ is the zero-based word index within $\boldsymbol{u}$. The final utterance embedding is the component-wise majority vote ($V$) of all of its position-encoded word vectors ($\boldsymbol{w_j^{pos}}$): 
\begin{align}
    \boldsymbol{w_j^{pos}} = \Pi^{l - j - 1}(\boldsymbol{w_j}) \\
    \boldsymbol{u} = V(\boldsymbol{w_0^{pos}}, \boldsymbol{w_1^{pos}}, ...  \boldsymbol{w_{l-1}^{pos}})
\end{align} 
Crucially, the generated sentence embedding is similar (as measured by cosine distance) to the embedding of all constituent words but maximally dissimilar to those of all others.

\subsection{Summary extraction}
Based on the obtained sentence embeddings, we identify the sentences that are the most representative of the document. We hypothesize that, if the obtained sentence embeddings sufficiently capture lexical and semantic characteristics of each sentence, the most important sentences will tend to appear at the center of sentence clusters, e.g. as medoids. Under such propositions,, we apply $k$-medoids algorithms to extract most central sentence embeddings as extractive summaries.
We apply \textit{alternating $k$-medoids} (the simplest $k$-medoid algorithm) to extract $k$ sentence embeddings from the pool of all sentence embeddings obtained from a document.

\begin{table*}[h]
 \centering
 \begin{tabular}{l|ccc|ccc|ccc|ccc}
 \toprule
     \multirow{2}{*}{\textbf{Model}}& \multicolumn{3}{c|}{\textbf{AMI}} & \multicolumn{3}{c|}{\textbf{ICSI}} & \multicolumn{3}{c|}{\textbf{Behance}}& \multicolumn{3}{c}{\textbf{ELITR}} \\
     & \multicolumn{1}{c}{\textit{R1}} & \multicolumn{1}{c}{\textit{R2}} & \multicolumn{1}{c|}{\textit{RL}} & \multicolumn{1}{c}{\textit{R1}} & \multicolumn{1}{c}{\textit{R2}} & \multicolumn{1}{c|}{\textit{RL}} & \multicolumn{1}{c}{\textit{R1}} & \multicolumn{1}{c}{\textit{R2}} & \multicolumn{1}{c|}{\textit{RL}} & \multicolumn{1}{c}{\textit{R1}} & \multicolumn{1}{c}{\textit{R2}} & \multicolumn{1}{c}{\textit{RL}}\\
    \midrule
    Lead-$n$ \cite{zhu2020make} & 0.60 & 0.40 & 0.60 & 0.44 & 0.19 & 0.44 & 0.31 & 0.18 & 0.31 & 0.36 & 0.14 & 0.36 \\
    TextRank \cite{mihalcea2004textrank} & 0.51 & 0.50 & 0.51 & 0.27 & 0.26 & 0.27 & 0.29 & \textbf{0.28} & 0.29 & 0.10 & 0.10 & 0.10 \\
    PacSum \cite{zheng-lapata-2019-sentence} & 0.05 & 0.04 & 0.05 & 0.06 & 0.03 & 0.06 & 0.32 & 0.20 & 0.32 & 0.14 & 0.03 & 0.14 \\
    OTExtSum \cite{tang-etal-2022-otextsum} & \textbf{0.66} & \textbf{0.41} & \textbf{0.66} & 0.585 & 0.28 & 0.585 & 0.31 & 0.16 & 0.37 & 0.45 & 0.17 & 0.45 \\
    HyperSum (ours) & 0.63 & 0.39 & 0.63 & \textbf{0.587} & \textbf{0.31} & \textbf{0.587} & \textbf{0.38} & 0.22 & \textbf{0.38} & \textbf{0.54} & \textbf{0.24} & \textbf{0.54} \\
    \bottomrule
 \end{tabular}
 \caption{Summary accuracy. All results are averages over 5 runs with different random seeds.}
 \label{tab:rouge}
\end{table*}

\textit{Alternating $k$-medoids} is a $k$-means like algorithm that discovers medoids iteratively, as described below:
\begin{enumerate}
    \item Randomly initialize $k$ medoids $\boldsymbol{\mu_1}, \boldsymbol{\mu_2}, ..., \boldsymbol{\mu_k} \in U$, where $U$ is the set of all sentence embeddings from the document to summarize.
    \item Until convergence, 
    \begin{enumerate}
    \item For each sentence embedding $u_m$, assign a label $c_m$ according to the nearest medoid.
    \begin{align}
    c_m:=\arg \min _n\left\|\boldsymbol{u_m}-\boldsymbol{\mu_n}\right\|^2
    \end{align}
    \item Update the medoids list (indexed with $n$) based on the distance to every sentence embedding.
    \begin{align}
    \boldsymbol{\mu_n}:=\frac{\sum_{i=1}^{|S|} 1\left\{c_m=n\right\} \boldsymbol{u_m}}{\sum_{m=1}^{|S|} 1\left\{c_m=n\right\}},\\
    \text{where $|S|$ is the total sentence count.} \nonumber
    \end{align}
    \end{enumerate}
\end{enumerate}

\section{Experiments and results}
\subsection{Datasets and metrics}
We assess the performance of HyperSum in four dialogue summarization benchmarks: AMI \cite{kraaij2005ami}, ICSI \cite{janin2003icsi}, Behance \cite{cho-etal-2021-streamhover}, and ELITR \cite{nedoluzhko-etal-2022-elitr}. All datasets except Behance are meeting transcripts. Behance consists of Internet livestream dialogues. Table \ref{tab:datasets} includes detailed statistics of each dataset. 

\begin{table}[th]
  \centering
  \begin{tabular}{cccc}
    \toprule
    \textbf{Name} & \textbf{Utts/Sample} & \textbf{Words/Utt} & \textbf{Ext. ratio} \\
    \midrule
    AMI & 845.35 & 7.51 & 0.34 \\
    ICSI & 1235.0 & 9.56 & 0.15\\ 
    Behance & 62.44 & 10.08 & 0.18 \\ 
    ELITR & 698.75 & 9.11 & 0.10 \\
    \bottomrule
  \end{tabular}
    \caption{Summary of benchmark datasets. "Utts/Sample", "Words/Utt", and "Ext. ratio" respectively denote average utterance count per document, average number of words per utterance, and the average ratio of reference summary utterance count to document utterance count.}
\label{tab:datasets}
\end{table}

We evaluate HyperSum with three metrics: overlap between generated and reference summaries, the faithfulness of summaries to the original document, and time elapsed during summary execution. We use ROUGE-1, ROUGE-2, and ROUGE-L \cite{lin-2004-rouge} as the degree of agreement between generated and reference summaries. We use the recently proposed ExtEval \cite{zhang-etal-2023-extractive} metric to measure the faithfulness of generated extractive summaries. ExtEval detects discourse-level utterance components, and determines if generated summaries include any incomplete or wrongly referenced discourse elements.

\subsection{Summarization accuracy}
ROUGE scores of baselines and HyperSum across all datasets are displayed in Table \ref{tab:rouge}. HyperSum outperforms state-of-the-art baselines in all datasets except AMI. Reported scores are generally higher than those reported by respective baselines because we provide the oracle number of summary sentences to extract, in the interest of fair comparison.

\subsection{Summarization faithfulness}

\begin{table}[h]
  \centering
  \begin{tabular}{l||c|c|c|c}
    \toprule
    \textbf{Model} & \textbf{AMI} & \textbf{ICSI} & \textbf{Behance} & \textbf{ELITR} \\
    \midrule
    Lead-$n$ \cite{zhu2020make} & 16.2 & 4.67 & 0.05 & 3.0\\
    TextRank \cite{mihalcea2004textrank} & 16.2 & 3.83 & 0.01 & 0.0 \\
    PacSum \cite{zheng-lapata-2019-sentence} & 1.9 & 0.17 & 0.04 & 0.0 \\
    OTExtSum \cite{tang-etal-2022-otextsum} & \textbf{17.7} & \textbf{5.17} & 0.06 & 2.0 \\
    HyperSum (ours) & 16.3 & 4.83 & \textbf{0.11} & \textbf{4.5} \\
    \bottomrule
  \end{tabular}
  \caption{Summary faithfulness.}
  \label{tab:faithfulness}
\end{table}

Even without any linguistic pretraining, HyperSum remains competitive with neural baselines in terms of summary faithfulness (Table \ref{tab:faithfulness}). HDC's amphibiousity that incorporates both distributional semantics and lexical representations prevents HyperSum from deviating too much from the original document. The learning-free nature of HyperSum also mitigates possible biases present in pretrained language models.

\begin{table*}[th]
  \centering
  \begin{tabular}{l||c|c|c|c||c|c}
    \toprule
    \textbf{Model} & \textbf{AMI} & \textbf{ICSI} & \textbf{Behance} & \textbf{ELITR} & \textbf{Mean} & \textbf{$\times$ HyperSum}\\
    \midrule
    TextRank \cite{mihalcea2004textrank} & 2.24 & 0.26 & 0.04 & 3.42 & 1.49 & $\times$ 2.81 \\
    PacSum \cite{zheng-lapata-2019-sentence} & 951.84 & 2288.49 & 4.62 & 720.00 & 991.24 & $\times$ 1870.26 \\
    OTExtSum \cite{tang-etal-2022-otextsum} & 16.81 & 26.36 & 7.27 & 21.91 & 18.09 & $\times$ 34.13 \\
    HyperSum (ours) & 0.39 & 1.19 & 0.05 & 0.51 & 0.53 & $\times$ 1 \\
    \bottomrule
  \end{tabular}
  \caption{Average seconds elapsed per utterance.}
  \label{tab:time_elapsed}
\end{table*}

\subsection{Summarization execution time}
On average, HyperSum is 2.8 times faster than the next fastest baseline, and 34 times faster than the fastest neural baseline (Table \ref{tab:time_elapsed}). HyperSum also shows the least amount of variance in running time across different benchmarks. Baselines' such as TextRank's and PacSum's runtime increases drastically with the average length of utterances in documents, due to their dependence on sentence graph construction. HyperSum avoids such structural bottlenecks by balancing fast initialization of random hypervectors with minor inaccuracies in their pseudo-orthogonality.

\subsection{Ablations}

We measure the effect of alternate $k$-medoid algorithms (Table \ref{tab:ablation_medoid}), HDC embedding methods (Table \ref{tab:ablation_embedding}), and utterance tokenization schemes (Table \ref{tab:ablation_tokenization}). All reported scores are ROUGE-L. In each table, the top row is the experiment configuration reported in Table \ref{tab:rouge}. Surprisingly, we found that using more modern $k$-medoids algorithms (that claim to improve upon naive \textit{alternating $k$-medoids} in accuracy) do not have meaningful impact in summary performance. 

\begin{table}[th]
  \centering
  \begin{tabular}{l||c|c|c|c}
    \toprule
    \textbf{Medoid Algo.} & \textbf{AMI} & \textbf{ICSI} & \textbf{Behance} & \textbf{ELITR} \\
    \midrule
    Naive (alternating) & \textbf{0.63} & \textbf{0.59} & \textbf{0.38} & \textbf{0.54} \\
    FastMSC \cite{schubert2022fast} &  0.59 & 0.52 & 0.27 & 0.44 \\
    FasterMSC \cite{schubert2022fast} & 0.61 & 0.43 & 0.27 & 0.43  \\
    FastPAM \cite{schubert2022fast}  & 0.62 & 0.58 & \textbf{0.38} & 0.51  \\
    FasterPAM \cite{schubert2022fast} & 0.62 & 0.58 & 0.37 & 0.51 \\
    \bottomrule
  \end{tabular}
  \caption{Effect of different $k$-medoid algorithms.}
  \label{tab:ablation_medoid}
\end{table}

Embedding words with thermometer HDC embeddings generate the best summaries except in AMI. One disadvantage of thermometer embeddings is in its vocabulary restriction. Thermometer embeddings in $D$ dimensions are limited to $D$ vocabularies, while normal hypervectors can represent $2^D$ symbols. This limitation in vocabulary size can be a disadvantage in situations where long documents must be summarized in small embedding units such as 3-grams. Thermometer embeddings will simply run out of encoding space. Segmenting utterances to word units for hyperdimensional encoding consistently performs the best. 

\begin{table}[th]
  \centering
  \begin{tabular}{l||c|c|c|c}
    \toprule
    \textbf{Embedding} & \textbf{AMI} & \textbf{ICSI} & \textbf{Behance} & \textbf{ELITR} \\
    \midrule
    Thermometer \cite{thermometer} & 0.63 & \textbf{0.59} & \textbf{0.38} & \textbf{0.54} \\
    Random \cite{kleyko2023survey} & 0.74 & 0.47 & 0.25 & 0.36 \\
    Level \cite{kleyko2023survey} & \textbf{0.75} & 0.58 & 0.26 & 0.48 \\
    Circular \cite{kleyko2023survey}  & 0.74 & 0.57 & 0.24 & 0.46 \\
    \bottomrule
  \end{tabular}
  \caption{Effect of different HDC embedding schemes.}
  \label{tab:ablation_embedding}
\end{table}

\begin{table}[th]
  \centering
  \begin{tabular}{l||c|c|c|c}
    \toprule
    \textbf{Tokenization} & \textbf{AMI} & \textbf{ICSI} & \textbf{Behance} & \textbf{ELITR} \\
    \midrule
    Word & \textbf{0.63} & \textbf{0.59} & \textbf{0.38} & \textbf{0.54}  \\
    Part-of-Speech & 0.61 & 0.51 & 0.29 & 0.45 \\
    SentencePiece \cite{kudo2018sentencepiece}  & 0.61 & 0.50 & 0.34 & 0.42  \\
    \bottomrule
  \end{tabular}
  \caption{Effect of different tokenization schemes.}
  \label{tab:ablation_tokenization}
\end{table}

\vfill\pagebreak

\section{Conclusion}

We present HyperSum with the aim of constructing an extractive summarizer that captures both efficiency and accuracy. Experiments across various benchmarks substantiate HyperSum's capability to improve upon both. We hope that HyperSum becomes a strong baseline for future extractive summarization research, especially in the unsupervised domain.

It is important to note that the two components of HyperSum -- \textit{sentence embedding generation} and \textit{extractive summarization through medoid selection} -- are orthogonal operations. Parallel research for improvement can be conducted in the two avenues. Similar to the way TextRank spun off a series of related research that either iterated upon its node construction or its node selection, HyperSum also leaves room for future research to iterate upon its two components. 

\label{sec:refs}

\bibliographystyle{IEEEbib}
\bibliography{strings,refs}

\end{document}